\documentclass[runningheads]{llncs}

\usepackage{graphicx}
\usepackage{color}
\usepackage{xcolor}
\usepackage{framed}
\usepackage{multirow}
\usepackage{amsmath}
\usepackage[caption=false]{subfig}
\usepackage{amssymb}
\usepackage{pifont}
\usepackage{hyperref}

\newcommand{\cmark}{\ding{51}}%
\newcommand{\xmark}{\ding{55}}%

\begin{document}

\newcommand{\sys}{Tab2Know}
\newcommand{\leanparagraph}[1]{\vspace{1mm}\noindent\textbf{#1.}}
\newcommand{\todo}[1]{{ \color{red} #1}}
\newcommand{\nary}{n-ary}

\newcommand{\conditional}[1]{{#1}}

\newcommand{\cX}[0]{\ensuremath{\mathcal{X}}}
\newcommand{\cY}[0]{\ensuremath{\mathcal{Y}}}

\newcommand{\cC}[0]{\ensuremath{\mathcal{C}}}
\newcommand{\cV}[0]{\ensuremath{\mathcal{V}}}
\newcommand{\cP}[0]{\ensuremath{\mathcal{P}}}
\newcommand{\cN}[0]{\ensuremath{\mathcal{N}}}
\newcommand{\fact}[1]{\ensuremath{\texttt{#1}}}
\newcommand{\constant}[1]{\ensuremath{\texttt{#1}}}

\newcommand{\squishlist}{
 \begin{list}{$\bullet$}
  { \setlength{\itemsep}{0pt}
     \setlength{\parsep}{3pt}
     \setlength{\topsep}{3pt}
     \setlength{\partopsep}{0pt}
     \setlength{\leftmargin}{1em}
     \setlength{\labelwidth}{1em}
     \setlength{\labelsep}{0.5em} } }

\newcommand{\squishend}{
\end{list}}

\newcommand{\rdfbegin}[0]{

\begin{center}
}

\newcommand{\rdfend}[0]{
\end{center}
}

\newcommand{\triple}[3]{
    \footnotesize
    \ensuremath{\texttt{#1}\;\;\texttt{#2}\;\;\texttt{#3}}
}

\newcommand{\triplel}[3]{
    \footnotesize
    \ensuremath{\texttt{#1}\;\;\texttt{#2}\;\;\texttt{"#3"}}
}

\newcommand{\triplei}[3]{
    \footnotesize
    \ensuremath{\texttt{#1}\;\;\texttt{#2}\;\;\texttt{#3\textasciicircum\textasciicircum}\langle\texttt{xsd:int}\rangle}
}

\author{Benno Kruit\inst{1,2} \and Hongyu He\inst{1} \and Jacopo Urbani\inst{1}}
\authorrunning{Benno Kruit \and Hongyu He \and Jacopo Urbani}
\institute{Department of Computer Science,\\ Vrije Universiteit Amsterdam, Amsterdam, The
Netherlands\\ \and Centrum Wiskunde \& Informatica, Amsterdam, The Netherlands \email{b.b.kruit@cwi.nl,hongyu.he@vu.nl,jacopo@cs.vu.nl}}

\title{\sys: Building a Knowledge Base \\ from Tables in Scientific Papers}

\maketitle

\begin{abstract} Tables in scientific papers contain a wealth of valuable
    knowledge for the scientific enterprise. To help the many of us who
    frequently consult this type of knowledge, we present \sys{}, a new
    end-to-end system to build a Knowledge Base (KB) from
    tables in scientific papers. \sys{} addresses the challenge of automatically
    interpreting the tables in papers and of disambiguating the entities that they
    contain. To solve these problems, we propose a pipeline that employs both
    statistical-based classifiers and logic-based reasoning. First, our pipeline
    applies weakly supervised classifiers to recognize the type of tables and
    columns, with the help of a data labeling system and an ontology
    specifically designed for our purpose. Then, logic-based reasoning is used
    to link equivalent entities (via \emph{sameAs} links) in different tables. An
    empirical evaluation of our approach using a corpus of papers in the Computer Science
    domain has returned satisfactory performance. This suggests that ours is a
    promising step to create a large-scale KB of scientific knowledge.

\end{abstract}

\section{Introduction}

Often, scientific advancement requires an extensive analysis of pre-existing
techniques or a careful comparison with previous experimental results. For
instance, it is common for researchers in Artificial Intelligence (AI) to ask
questions like ``Which are the most popular datasets used for graph
embeddings?'' or ``What is the F1 of BERT on TACRED?". Finding the answers
obliges the researchers to spend much time in perusing existing
literature, looking for experimental results, techniques, or other valuable
resources.

The answers to such questions can be frequently found in tabular form,
especially the ones that describe the output of experiments. Unfortunately,
tables in papers are made for human consumption; thus, their layout can
be irregular or contain specific abbreviations that are hard to disambiguate
automatically. It would be very useful if their content were copied into a clean
Knowledge Base (KB) where tables are disambiguated and connected using a single
standardized vocabulary. This KB could assist the users in finding those answers
without accessing the papers or could be used for many other purposes,
like categorizing papers, finding inconsistencies or plagiarized content.

To build such a KB, we present \sys{}, an end-to-end system
designed to interpret the tables in scientific papers. The main challenge
tackled by \sys{} lies in the interpretation of the table, which is a necessary
step to build a KB. In this context, the peculiarities of tables in scientific
literature make our domain quite different from previous work
(e.g.,~\cite{tabel,ritze_matching_2015,kruit_extracting_2019}), which mainly
focused on Web tables. First, the interpretation of Web tables benefits from the
existence of large, curated KBs (e.g., DBPedia~\cite{dbpedia}), which allows the
linking of many entities. In our case, there is no such KB. Second, a large
number of Web tables can be categorized as \emph{entity-attribute} tables, i.e.,
tables where each row describes one entity, and the columns represent
attributes~\cite{kruit_extracting_2019,ritze_matching_2015,zhang2017effective}.
In our context, we observed that many tables are of different types, namely they
express n-ary relations, such as the results of experiments. For such tables,
existing techniques designed for entity-attribute tables cannot be reused.

With \sys{}, we propose a pipeline for knowledge extraction that includes both
weakly supervised learning methods and logical reasoning. \sys{} is designed to
1) detect the type of the table; 2) disambiguate the types of columns, and 3)
link the entities between tables. The first operation is applied to distinguish, for
instance, tables that report experiments from tables that report examples. The
second operation recognizes the rows that contain the headers of the table and
disambiguates the columns, linking them to classes of an ontology. The third
operation links entities in different tables.

We implement the first two operations using statistical-based classifiers
trained with bag-of-words and context-based features. These classifiers have
an accuracy that largely depends on the quality and amount of training data.
Unfortunately, labeling training data is increasingly the largest bottleneck as
it often requires an expensive manual effort and/or expertise that might not
be readily available. To counter this problem, we propose a weakly supervised
method that relies on SPARQL queries and Snorkel~\cite{snorkel}. The SPARQL queries
are used to automatically retrieve samples of a given class,
type, etc., while Snorkel resolves potential conflicts in the prediction
with a sophisticated voting mechanism.

After the first two operations are completed, we transform the tables into an RDF
KB and apply reasoning with existentially quantified rules to identify and link
entities in different tables. Reasoning with existentially quantified rules is a
well-known technology for data integration and wrangling~\cite{rules_wrangling}.
For our problem, we designed a set of rules that considers the types of columns
and string similarities to establish links using the \emph{sameAs} relation.
Then, we used VLog~\cite{vlog} to materialize the derivations and link the
entities across the tables.

We evaluated our approach considering open access CS
papers. In particular, we evaluated the performance of our
pipeline using gold standards and compared it to another state-of-the-art method.
We also applied our method to a larger corpus with 73k scientific tables.
In these tables, we found 312k entities, which are linked to the table structure
and metadata in our large-scale KB.

\conditional{We release the datasets, gold standards, and resulting KB as an open
resource for the research community at
\url{https://doi.org/10.5281/zenodo.3983012}. The code, ruleset, and
instructions to replicate our experiments are also publicly
available at \url{https://github.com/karmaresearch/tab2know}.}


\section{Related Work}
\label{sec:relatedwork}

Extracting knowledge from tables is a process that can be divided into
\emph{three} main tasks: \emph{table extraction}, \emph{structure detection},
and \emph{table interpretation}. \conditional{Once a set of tables is interpreted, another
problem consists of recognizing whether multiple tables mention the same
entities. We call this task \emph{entity linking}, but this is also known as
\emph{entity resolution}~\cite{papadakis2020entity}, \emph{record
    linkage}~\cite{christen_survey_2012}, or \emph{entity
matching}~\cite{bohm_linda_2012}.}


\leanparagraph{Table extraction} This task consists of recognizing the parts of
a PDF/image which contain a table. 
Existing methods can be categorized either as
heuristic (e.g.,~\cite{trex,pdffigures2}) or supervised
(e.g.,~\cite{mccallum_extract}). 
%
In this paper, we use the system \emph{PDFFigures}~\cite{pdffigures2}, which is
a recent approach based on heuristics with very high precision and recall ($\geq
90\%$) that is used in Semantic Scholar~\cite{ammar:18}.

\leanparagraph{Structure detection} Given as input an image-like representation
of a table, some systems focus on recognizing the table's structure so that it
can be correctly extracted. A popular system is \emph{Tabula} (\url{
    https://tabula.technology/}),
which recognizes the table's structure using rules. More recently, some deep learning methods based on Convolutional Neural
Networks (CNN)~\cite{schreiber_deepdesrt_2017}, Conditional Generative Adversarial Networks (CGAN)~\cite{cgan}, and a combination of a CNN, saliency and graphical
models~\cite{kavasidis_saliency-based_2019} have been evaluated. The performance of these methods
is good ($F_1\geq 0.95$), but not much different from Tabula, which
returns a $F_1$ between 0.86 and 0.96 and has the advantage that is unsupervised.

\leanparagraph{Table interpretation} The goal of table interpretation consists
of linking the content of the table to a KB so that new knowledge can be
extracted from the table~\cite{limaye_annotating_2010}. In this context, most of
the previous work has focused on tables that represent \emph{entity-attribute}
relations~\cite{kruit_extracting_2019}. These tables have rows that describe
entities and columns that describe attributes. Thus, their interpretation
consists of mapping each row to an entity in the KB, and linking each column to
a relation in the KB. Some work has focused only on the first task
(e.g.,~\cite{tabel}) while others on the second
(e.g.,~\cite{chen_colnet_2019,efthymiou_matching_2017,LUOLCZ18}).
\conditional{The work at~\cite{chen_colnet_2019}, in particular, is similar to
    ours as it also uses SPARQL queries to create training data.
    The difference is that in~\cite{chen_colnet_2019}, SPARQL is used to query a rich KB automatically, whereas in our case, we let users specify queries since we lack such a KB.}
In terms of methodology, current work in this field either relies on statistical
models, like PGMs~\cite{limaye_annotating_2010,tabel}, or introduces an
iterative process that filters out
candidates~\cite{ritze_matching_2015,ritze_profiling_2016,zhang2017effective}.

As far as we know, the only systems that offer a end-to-end table
interpretation are \emph{T2K}~\cite{ritze_matching_2015},
\emph{TableMiner+}~\cite{zhang2017effective}, and
\emph{TAKCO}~\cite{kruit_extracting_2019}, but these are designed for
Web tables and rely on a rich KB like DBPedia~\cite{dbpedia}, which we do not
have.

The only work that has focused on the interpretation of tables from scientific
literature is~\cite{tables_www2020}. The authors describe an
approach to automatically extract experimental data from tables based on ensemble learning. Although we
view this work as the most relevant to our problem, there are several important
differences between our work and theirs. First, our approach employs a different
set of technologies and performs entity linking, which is not considered
in~\cite{tables_www2020}. Then, our approach is more general. In
fact, \cite{tables_www2020} focuses only on the extraction of tuples
$(method,dataset,metric,score,source)$ while ours extracts a larger variety of
knowledge. Finally, our approach yields a better accuracy
(see Section~\ref{sec:evaluation}).

\conditional{ \leanparagraph{Entity Linking} The problem of resolving entities
    in tables has received considerable attention in database research (96+
    papers in VLDB, KDD, etc. in
    2009-2014)~\cite{papadakis2020entity,christen_survey_2012,konda_magellan_2016}.
    One of the most popular systems is Magellan~\cite{konda_magellan_2016}.
    Magellan is a tool to help users to perform entity matching, providing
    different implementations of matching and blocking algorithms. Recently,
    Mudgal et al.~\cite{mudgal_deep_2018} have studied the application of deep
    learning for entity matching, but concluded that it does not outperform
    existing methods on structured data. Other works have explored the usage of
    embeddings for this task: For instance, Cappuzzo et
    al.~\cite{cappuzzo_creating_2020} have shown how we can construct embeddings
    from tabular data. Another line of work has been focusing on crowds,
    e.g.,~\cite{das_falcon_2017} and citations therein, while other works have
    focused on entity resolution using knowledge bases (e.g.,
    LINDA~\cite{bohm_linda_2012}). Our work differs from the ones above because
    they either focus on highly structured table sets or require the existence of
    KBs (which we do not have). Moreover, another important difference is that we take
    a declarative approach with rules. Rules are useful because they can
    be easily debugged/extended directly by domain experts, and they can be
    integrated with ontological reasoning.

}

\leanparagraph{Other related works} We mention, as further related work, the
systems by~\cite{findrelatedtables} and \emph{TableNet}~\cite{tablenet} which
focus on \emph{searching} for tables related to a given query. Other, less
relevant works focus on extracting and searching for figures on
papers~\cite{figureseer,figures3}. These works complement our approach and can
further assist the user to find relevant knowledge in papers.


\section{Overview}
\label{sec:overview}

\begin{figure}[t]
 \centering
 \includegraphics[width=0.95\textwidth]{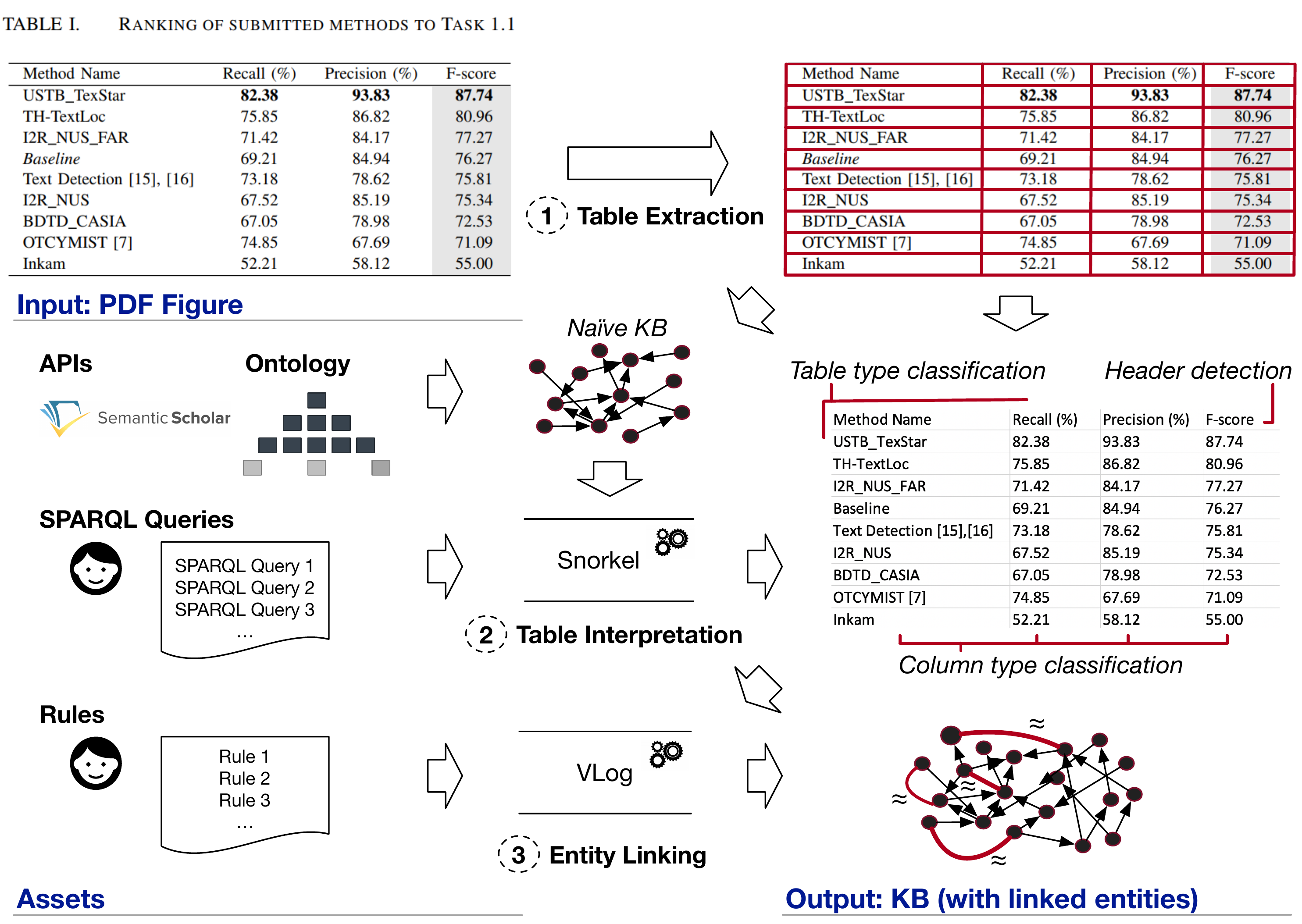}
 \caption{\sys: System Overview}
 \label{fig:overview}
\end{figure}

Our goal is to construct a clean and large KB from the content of tables in
scientific papers stored as PDFs. To do so, we need to address two main
challenges: first, we must resolve the ambiguities that might arise during the
noisy extraction process and reduce the error rate as much as possible. Second,
we must counter the problem that we lack both: 1) a pre-existing KB that can guide the
extraction process and 2) a large amount of training data. We must, in other
words, find a way to build a KB from scratch.

Our proposal is a pipeline with three main tasks, as shown in
Figure~\ref{fig:overview}:

\squishlist

\item \textbf{Task 1: Table Extraction.} The system receives as input an image-like
    representation of a table, recognizes its structure, and returns its content
    as a CSV file. For this task, we use external tools. We provide more details
    below;

\item \textbf{Task 2: Table Interpretation.} The system processes the CSV to recognize
    the headers and the type of the table. Then, it disambiguates the columns by
    mapping them to ontological classes. We describe this task in
    Section~\ref{sec:interpretation};

\item \textbf{Task 3: Entity Linking.} Finally, the
    system performs logical-based reasoning to link the entities across tables.
    We describe this task in Section~\ref{sec:entitylinking}.

\squishend

While in principle our method can be applied to scientific papers in any domain,
we restrict our analysis to papers in Computer Science, which is our area of
expertise. In particular, we consider Open Access papers and have been published
in top-tier venues in subfields like AI, semantic web, databases, etc.

Before we describe the components, we describe \emph{two} additional assets that
we use for different purposes. %
%
\conditional{The first one is an ontology constructed annotating a sample of
    random tables. A first version of this ontology contained 44 classes
    organized in a hierarchy with a maximum depth of 6. After further
    annotations, we decided to simplify it to a set of of 27 classes (depth 3)
    for which we had substantial evidence in our corpus. The final ontology has
    4 root classes: $\mathbf{Example}$, $\mathbf{Input}$,
    $\mathbf{Observation}$, and $\mathbf{Other}$. These classes define
    general table types. Then, the subclasses describe column types, e.g.,
    $\mathbf{Dataset}$, $\mathbf{Runtime}$, or $\mathbf{Mean}$. As an example,
    one of the longest chains is $\mathbf{Recall} \sqsubseteq
    \mathbf{Metric} \sqsubseteq \mathbf{Observation}$ with $\sqsubseteq$ denoting the subclass
relation. The ontology is serialized in OWL using WebProt\'eg\'e~\cite{protege}
and is publicly available as resource.}

The second asset is an external KB that contains metadata of the papers, namely
Semantic Scholar~\cite{ammar:18}. We access it using the provided APIs to
retrieve the list of authors, the venue, and other contextual data.

\leanparagraph{Table Extraction} Our input consists of a collection of papers in
PDF format. The first operation consists of launching
PDFFigures~\cite{pdffigures2} to extract from the PDFs the coordinates of tables
and related captions. We use the coordinates to extract an image-like
representation of the tables, see for instance the table reported in
Figure~\ref{fig:overview}.  Then, we invoke Tabula, which is a
tool also used in similar prior works~\cite{tables_www2020}, to recognize the
structure of the tables using their coordinates and to translate them into CSV
files.


After the images are converted, we perform a na\"ive conversion of the tables
into RDF triples. We assign a URI to every table, column, row, and cell and
link every cell, row, and column to the respective table with positional coordinates.

\begin{example}\label{ex:naivedump} Consider the table in
    Figure~\ref{fig:overview}. We report below some triples that are generated while dumping
    its content into RDF.
    \rdfbegin
    \begin{tabular}{l@{\hskip 0.02in}l}
    \multicolumn{2}{l}{\footnotesize\ensuremath{\texttt{PREFIX : http://xzy/tab2know}}} \\
    \triple{:Table1}{:hasRow}{:Table1-r1} & \triple{:Table1}{:hasCol}{:Table1-c1}\\
    \triple{:Table1-r1}{rdf:type}{:Row} & \triple{:Table1-c1}{rdf:type}{:Column}\\
    \triplei{:Table1-r1}{:rowIndex}{1} & \triplei{:Table1-c1}{:colIndex}{1}\\
    \triple{:Table1-r1c1}{:cellOf}{:Table1} & \triple{:Table1-r1c1}{rdf:type}{:Cell} \\
    \triplei{:Table1-r1c1}{:rowIdx}{1} & \triplei{:Table1-r1c1}{:colIdx}{1} \\
    \triplel{:Table1-r1c1}{rdf:value}{Method name} & \triplel{:Table1-r2c1}{rdf:value}{USTB\_TexStar}\\
    \multicolumn{2}{c}{\texttt{...}}
    \end{tabular}
    \rdfend
\end{example}

As we can see from the triples in Example~\ref{ex:naivedump}, the KB generated
at this stage is a direct conversion of the tabular structure into triples.
Despite its simplicity, however, such a KB is already useful because it can be
used to query the n-ary relations expressed in the tables in combination with
the papers' metadata. For instance, we can write a SPARQL query to retrieve all
the tables created by one author with a caption containing the word
``results'', or to retrieve the tables containing ``F1'' and which appear as proceedings of a
certain venue.

The main problem at this stage is that we can only query using string
similarities, which severely reduces the recall. For instance, a query could
miss a column titled \texttt{Prec.} if it searches for \texttt{Precision}. The
next operation, described below, attempts to disambiguate the tables to create
a KB that is more robust against the syntactic diversity of the surface form of
their content.


\section{Table Interpretation}
\label{sec:interpretation}

\sys{} performs three main operations to interpret the tables. First, it
identifies the rows with the table's header
(Section~\ref{sec:header}). Then, it detects the type of the table
(Section~\ref{sec:tabletype}). Finally, it maps each column to an ontological
class (Section~\ref{sec:columns}). First, we describe the procedure to obtain
training data.

\subsection{Training Data Generation} \label{sec:weak}

Statistical models are ideal for implementing a table interpretation that is robust
against noise. However, their accuracy depends on high-quality training data,
which we do not have (and it is expensive to obtain such data with human
annotators). We counter this problem following the paradigm of \emph{weak
supervision}. The idea is to employ many annotators, which are much
cheaper than a human expert but also much noisier. These annotators can deliver
a large volume of labeled data, but the labels might be incorrect or
conflicting. To resolve these problems, we can either rely on procedures like
majority voting or train a dedicated model to computed the most likely correct
label. In the second case, we can use Snorkel, one of the most popular models
for this purpose~\cite{snorkel}.

Snorkel's goal is to facilitate the learning of a model $\theta$ that, given a
data point $x \in \cX$, predicts its label $y \in \cY$. Instead of training
$\theta$ by fitting it to a set of pre-labeled data points, as it would happen
in a traditional supervised approach, Snorkel trains an additional generative
model with unlabeled data and uses pre-labeled data only for validation and
testing. For these two tasks, the amount of pre-labeled data can be much
smaller, and thus cheaper to obtain. Then, the generative model can be used to
train $\theta$.

Snorkel introduces the term \emph{labeling function} to indicate a data
annotator with possibly low accuracy. A labeling function $\lambda : \cX
\rightarrow \cY \cup \{\emptyset\}$ can encode a heuristic or be a simple
predictor. It receives a data point $x$ in input and either returns a label in
$\cY$ or \emph{abstains}, i.e., returns $\emptyset$. Given $m$ unlabeled data
points and $n$ labeling functions, Snorkel applies the labeling functions to
the data points and computes a matrix $M \in (\cY \cup \{\emptyset\})^{m\times n}$.

Then, Snorkel processes $M$ to compute, for each $x_i$ where
$i\in\{1,\ldots,m\}$, a \emph{probabilistic training label} $\tilde{y_i}$. The
processing consists of creating a generative model using a matrix
completion-style algorithm over the covariance matrix of the
labels~\cite{ratner2019training}. Then, this model can be used to generate
labeled data for training $\theta$. In this work, we considered models such as
Na\"ive Bayes (NB), Support Vector Machine (SVM), and Logistic Regression
(LR)~\cite{mlbook} to implement $\theta$. We have also experimented with deeper
learning models, but we did not obtain improvements because such models are more
prone to overfitting if training data is scarce.



The effectiveness of Snorkel largely depends on the number and quality of the
labeling functions. In our context, we implemented them using SPARQL queries,
which are supposed to be entered by a (human) user. SPARQL queries are ideal
because they can assign labels to many data points at once.
For each query $Q$, we create a labeling function that receives in input a
column/table $x$ and returns an assigned class label (e.g., a table type, or the
class of a column) if $x$ is among the answers of $Q$. Otherwise, the function
abstains.

\begin{example} We show below an example of a SPARQL query that labels columns
    with the class $\mathbf{F_1}$ if they have a header cell with value ``f1'' and
    contain any cell with a numeric type.

    \vspace{0.5em}
    \begin{minipage}{\linewidth}
    \footnotesize
    \hangindent=0.7cm
    \texttt{
select distinct ?column where \{  \\
\hspace*{2ex} ?table :column ?column ; :cell ?cell . \\
\hspace*{2ex} ?column :hasTitle "f1" . ?cell rdf:type xsd:decimal . \}}
\end{minipage}
\vspace{0.3em}

Clearly, this query is not a good predictor if taken alone, but if we combine its output with the ones of many other
functions, then the resulting predictive power is likely to be superior. This is
the key observation used by Snorkel.
\end{example}

In our pipeline, we execute all the user-provided SPARQL queries and then use
their outputs to build the matrix $M$ for a large number of data points. Next,
we train the final discriminative model $\theta$. We compute
two different $\theta$: One to generate training data for predicting the tables'
types (Section~\ref{sec:tabletype}) while the other is for predicting the
columns' types (Section~\ref{sec:columns}).


\subsection{Table Header Detection}\label{sec:header} First, we identify the
rows that define the headers. To this end, we can either always select the first
row as header or employ more sophisticated methods to recognize multi-row
headers, like~\cite{fang2012table}. We observed that a simplified unsupervised
version of~\cite{fang2012table} yields a good accuracy on our dataset. We
describe it below.

Our procedure exploits the observation that header rows differ significantly
from the rest of the table with respect to character-based statistics. Hence, we
categorize characters either as \emph{numeric}, \emph{uppercase},
\emph{lowercase}, \emph{space}, \emph{non-alphanumeric}, or \emph{other}. Then,
for each column, we count how many characters of each class (e.g., numeric)
appear in its cell. We compute the average count per class across the column and
use these values to determine the standard deviation for each cell. The
\emph{outlier score} of a row $r$ is determined as the average of the standard
deviations of all classes of its cells. If the
outlier score or $r$ is greater than $\tau$ (default value is $1$, set after
cross-validation), then $r$ is marked as header.

\subsection{Table Type Detection} \label{sec:tabletype}

\begin{figure}[t]
 \centering
 \includegraphics[width=\textwidth]{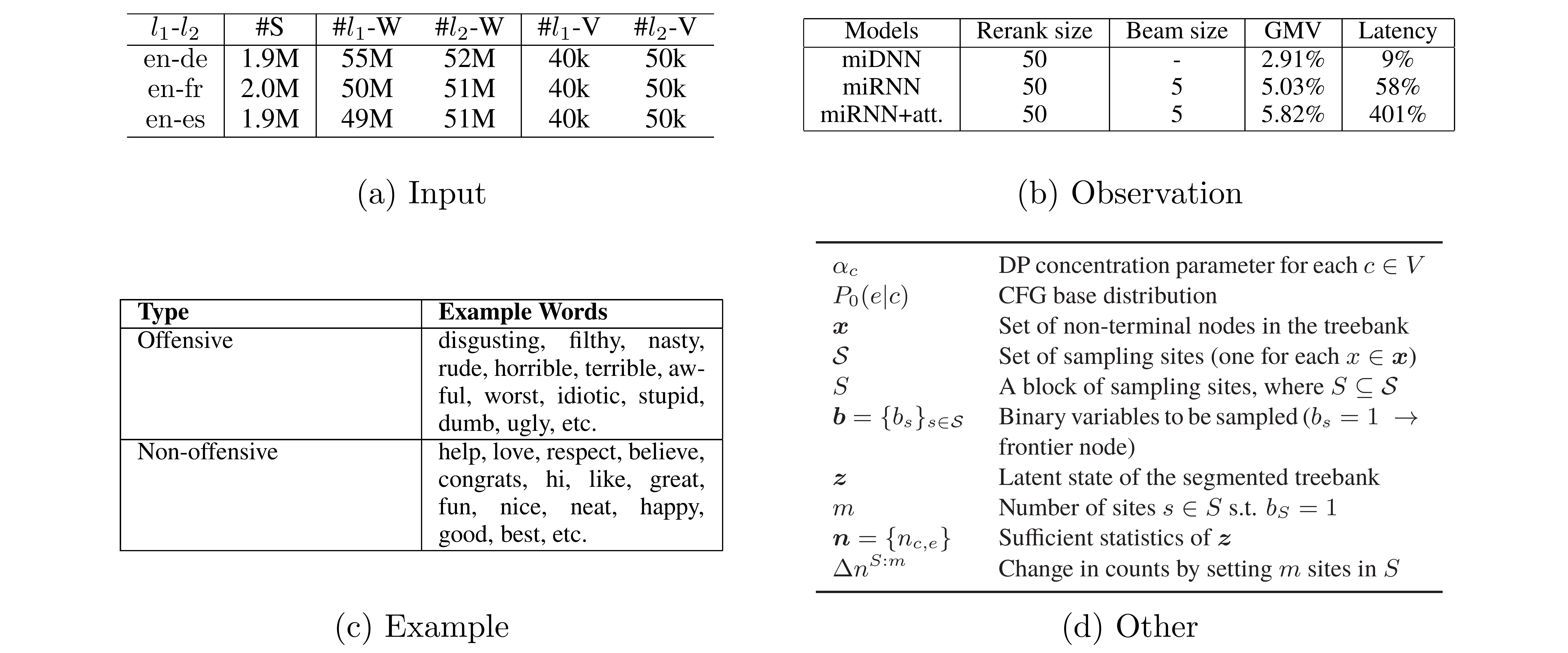}
 \caption{Examples of tables of each category}
 \label{fig:exampletables}
\end{figure}

In scientific papers, tables are used for various reasons. We classified them in
the classes $\mathbf{Observation}$, $\mathbf{Input}$, $\mathbf{Example}$,
and $\mathbf{Other}$ (See Fig.~\ref{fig:exampletables} for
examples).

Knowing the class of a table is useful for reducing the search space when the user
is interested in some specific content (e.g., The $F_1$ measure is typically not
mentioned in tables of type $\mathbf{Example}$). Moreover, we can also use this information as
a feature for the column disambiguation.


We predict the table type with a statistical classifier. As features for the
classifier, we selected bags-of-ngrams of lengths 1 to 3 that occurred more than
once, weighted by their TF-IDF score. Tables often contain abbreviations and
domain-specific symbols that address an audience of experts. These provide
strong hints for determining the type of the table; thus we consider the ngram
in the content of the cells and the table caption. We also included other
numerical features. In particular, we use the fraction of numeric cells in the
table and the minimum, maximum, median, mean and standard deviation of
numerical columns. This resulted in a total of 5804 features.

%
To train the models, we first ask the users to specify some SPARQL queries which
will be used by Snorkel to create a large volume of training data. Then, we
experimented with three well-known types of classifiers: NB, SVM, and LR.
Eventually, we selected LR because it returned the best performance on the
noisiest dataset.


\subsection{Column Type Detection} \label{sec:columns}

Finally, the interpretation procedure attempts at linking the columns to one of
the available classes in our ontology. The ontology includes popular classes
that we identified while annotating a sample (e.g., $\mathbf{Dataset}$,
$\mathbf{Runtime}$,\ldots), while infrequent classes with very few columns are
mapped to the class $\mathbf{Other}$. In general, we assume that a column is
\emph{untyped} if it is mapped to $\mathbf{Other}$.

For this task, we also used bag-of-ngram features of lengths 1 to 3, extracted
from the table caption, the column header cells, the header cells of the other
columns, and the column body. We restricted the set of ngrams to only the top
1000 most frequent per extraction source. Additionally, we added features about
the numerical columns, identical to those in Section~\ref{sec:tabletype}.  This
resulted in a total of 3076 features.

Similarly as before, we first rely on user-provided SPARQL queries to generate
training data. Then, we considered NB, SVM, and LR as classifiers. Once the
models for the table and column types are trained, we use them to predict the
types of every table and column in our corpus. Finally, we use the predicted class
to annotate the table/column in the KB with a semantic type.

\section{Entity Linking}
\label{sec:entitylinking}

\leanparagraph{Rationale} Predicting the types of tables and columns is useful
to map the table schema into a meaningful \nary{} relation. The last operation
in our pipeline consists of associating cells to entities so that we can
populate the \nary{} relations with new instances.

We start by assuming that every non-numerical cell contains an entity mention,
which implies the existence of one entity. This assumption is not unrealistic.
Indeed, if we look at the table in Figure~\ref{fig:overview}, then we see that
every non-numerical cell that is not in the table's header refers to an entity
(e.g., the cell ``USTB\_TextStar'' refers to an algorithm to detect text inside
images).


In practice, it is likely that some entities are mentioned multiple times.  This
consideration motivates us to discover whether two entity mentions (possibly on
different tables) refer to the same entity. When we do so, then we gain more
knowledge about the entity and reduce the number of entities in the target KB.
We call this task \emph{entity linking} because we are linking, with the
\emph{sameAs} relation, equivalent entities across tables.

With this goal in mind, we start by assuming that every entity has the content
of the corresponding cell as label. For instance, the entity mentioned in the
cell with ``USTB\_TextStar'' has ``USTB\_TextStar'' as label. Using the labels
to determine equality can be surprisingly effective in practice, but it is not
an operation without risks. In fact, there are cases where different entities
have the same label, or the same entity has multiple labels. These cases call
for a more sophisticated procedure to discover equalities.



\leanparagraph{Reasoning} Reasoning with existentially quantified rules is an
ideal tool to establish non-trivial equalities between entities since it was
already previously used for data integration problems~\cite{fagin,llunatic}.
For our purposes, we are interested in applying two types of rules: \emph{Tuple
Generating Dependencies (TGDs)} and \emph{Equality Generating Dependencies
(EGDs)}. We describe those below.

Consider a vocabulary consisting of infinite and mutually disjoint sets of
predicates $\cP$, constants $\cC$, null values $\cN$, and variables $\cV$. A
\emph{term} is either a constant, a variable, or a null value. An \emph{atom} is
an expression of the form $p(\vec{x})$ where $p\in\cP$, $\vec{x}$ is a tuple of
terms of length equal to the arity of $p$, which is fixed. A \emph{fact} is an
atom without variables. A TGD is a rule of the form: \begin{equation} \forall
    \vec{x},\vec{y}.(B\rightarrow \exists\vec{z}.H) \end{equation}

\noindent where $B$ is a conjunction of atoms over $\vec{x}$ and $\vec{y}$ while
$H$ is a conjunctions of atoms over $\vec{y}$ and $\vec{z}$. Let $x,y \in
\vec{x}$. A EGD is a rule of
the form: \begin{equation}
    \forall \vec{x}.(B\rightarrow x\approx y)
\end{equation}

Intuitively, TGDs are used to infer new facts
from an existing set of facts (i.e., the database). Their execution consists of
finding in the database suitable replacements for the variables in $\vec{x}$ and
$\vec{y}$ that render $B$ a set of facts in the database. Then, these
replacements and mappings from $\vec{z}$ to fresh values in $\cN$ are used to
map $H$ into a set of facts, which is the set of \emph{inferred} facts.

EGDs are used to establish the equivalence between terms. Their execution is
similar to the one of TGDs, with the difference that whenever they infer that $a
\approx b$, where $a$ and $b$ are terms and $a < b$ according to a predefined
ordering, then every occurrence of $b$ in the database is replaced with $a$.

The \emph{chase}~\cite{fagin} is a class of forward-chaining procedures that
exhaustively apply TGDs and EGDs to infer new knowledge with the rules. A formal
definition of various chase procedures is available at~\cite{benchmarking}. In
this work, we apply the restricted chase, one of the most popular variants. It
is known that sometimes the chase may not terminate, but this is not our
case since we use an \emph{acyclic} ruleset~\cite{fagin}.

We first map the content of the KB extracted from the tables
into a set of facts. For example, the first two RDF triples in
Example~\ref{ex:naivedump} map to the facts
$\fact{hasRow(Table1,Table1-r1)}$ and $\fact{hasCol(Table1,Table1-c1)}$
respectively.

Then, we use the two TGDs \begin{align} type(X,\constant{Column})\rightarrow
\exists Y. colEntity(X,Y) \tag{$r_1$} \\ type(X,\constant{Cell})\rightarrow
\exists Y. cellEntity(X,Y) \tag{$r_2$} \end{align}

\noindent to introduce fresh entities for every column and cell in the tables.
The predicates $colEntity$ and $cellEntity$ link entities ($Y$) to the columns
and cells respectively. Note that we use null values to represent entities, thus
we are simply stating their existence with some placeholders.  To reason and
discover whether two different entities are equivalent, we employ EGDs.  In
particular, we use five EGDs, reported below:
%
\begin{align} ceNoTypLabel(X,L),ceNoTypLabel(Y,L) \rightarrow X\approx Y
    \tag{$r_3$} \\ eNoTypLabel(X,C,L),eNoTypLabel(Y,C,L) \rightarrow X \approx Y
    \tag{$r_4$} \\ eTableLabel(X,T,L),eTableLabel(Y,T,L) \rightarrow X \approx Y
    \tag{$r_5$} \\ eTypLabel(X,S,L),eTypLabel(Y,S,M),STR\_EQ(L,M) \rightarrow X
\approx Y \tag{$r_6$} \\ eAuthLabel(X,A,L),eAuthLabel(Y,A,M),STR\_EQ(L,M)
\rightarrow X \approx Y \tag{$r_7$} \end{align}

\noindent where $ceNoTypLabel$, $eNoTypLabel$, $eTableLabel$, $eTypLabel$, and
$eAuthLabel$ are auxiliary predicates that we introduce for improving the
readability. We describe their intended meaning as follows. The fact
$ceNoTypeLabel(X,L)$ is true if $colEntity(Y,X)$ is true and $Y$ is an untyped
column with header value $L$; $eNoTypeLabel(X,C,L)$ is true if $X$ is an entity
with a label $L$ that appears in a cell inside an untyped column associated to
entity $C$; $eTableLabel(X,T,L)$ is true if entity $X$ with label $L$ appears in
table $T$; $eTypeLabel(X,S,L)$ is true if entity $X$ with label $L$ appears in a
column with type $S$; $eAuthLabel(X,A,L)$ is true if entity $X$ with label $L$
appears in a table authored by author $A$.

The rationale behind each EGD is the following:
\squishlist

\item $\mathbf{Rule\; r_3:}$ This rule is introduced to disambiguate untyped
    columns. Since we were unable to discover the columns' types and assigned
    them to the class \texttt{Other}, we use the value of the header to
    determine whether they contain the same type of entities. Thus, the rule
    will infer that their associated entities are equal if they share the same
    header.

\item $\mathbf{Rule\; r_4:}$ This rule infers that two entities are equal if
    they appear in the same group of columns (created by $r_3$), and
    they share the same label.

\item $\mathbf{Rule\; r_5:}$ This rule encodes a simple heuristics, namely that
    if two entities with the same label appear in the same table, then they
    should be equal, irrespective of the type of columns where they appear.

\item $\mathbf{Rule\; r_6:}$ This rule disambiguates entities in columns of the
    same type. Here, we no longer consider the header of the column (as done by
    $r_3$ and $r_4$) but compare the entities' labels. After experimenting with
    approximate string similarity measures, like the Levenshtein distance, we
    decided to use a case insensitive string equality ($STR\_EQ$) to reduce the
    number of false positives. Case-insensitive similarity is more expensive
    than an exact string match because it requires dictionary lookups. We use it
    here and not in $r_3$, $r_4$, and $r_5$ because the comparisons are
    done only between entities of the same type.

\item $\mathbf{Rule\; r_7:}$ This rule implements another heuristic which takes
into account the authors of the paper. It assumes that two entities are equal if
they appear in two tables authored by the same author (we used the
IDs provided by Semantic Scholar to disambiguate authors) and have the same label.

\squishend

Once the reasoning has terminated, we introduce a new entity for each different
null value and add RDF triples that link them to the corresponding cells and
columns. Notice that the list of presented rules is not meant to be exhaustive.
The ones that we describe show how we can exploit the predictions computed in
the previous step ($r_6$) and external knowledge ($r_7$) relying on string
similarity when no extra knowledge is available. We believe that additional
EGDs, possibly designed to capture some specific cases, can further improve the
performance.

\section{Evaluation}
\label{sec:evaluation}



\leanparagraph{Inputs} We considered two datasets: A corpus of tables that we
manually constructed, and the dataset by~\cite{tables_www2020}, which is called
\emph{Tablepedia}.

Our corpus of tables contains 142,966 open-access PDFs distributed by Semantic
Scholar. These papers appear in the proceedings of top venues in CS (the  full list of
venues is reported in our data repository). From these papers, we extracted
73,236 tables with PDFFigures and Tabula.
\conditional{
    These tables have 6.23 rows on average (SD = 6.58), and they have 7.11 columns (SD = 6.27).
}
We converted the tables into RDF,
resulting in a KB with 23M triples. We used Blazegraph
 to
execute the SPARQL queries. After adding the table types and column types, we
loaded the KB into VLog~\cite{vlog} to perform rule-based reasoning.

Tablepedia contains 451 tables, which have the columns annotated
only with three classes: $\mathbf{Method}$, $\mathbf{Dataset}$, and $\mathbf{Metric}$.  To
use this dataset in our pipeline, we created a graph representation of the
tables without the annotations. Then, we translate the 15 \emph{seed concepts}
that are used in~\cite{tables_www2020} to create the tables into labelling
queries, so that we could apply Snorkel using both datasets.  \conditional{ In
    contrast to Tablepedia, our annotated dataset maps to a much larger number
    of classes. Notice that the most frequent column
    types
    in our dataset ($\mathbf{Observation}$, $\mathbf{Accuracy}$, and $\mathbf{Count}$),
    do not occur in Tablepedia.  }

\conditional{ \leanparagraph{Training data} To create the training data for weak
    supervision, two human annotators (one PhD and one bachelor CS student)
    wrote SPARQL queries for labeling with the aid of a web interface designed
    for this purpose. The
    annotators examined the results of these queries on a sample of 400 tables,
    ensuring that the queries represented heuristics that covered a reasonable
    amount of the data. The quality of the SPARQL queries is fundamental to
    produce a good training dataset, and hence return good predictions. It is
    crucial that the queries have \emph{large coverage} to avoid introducing a
    bias and to increase the training data size. For instance, if the queries
    label only a few tables, then the model will not receive enough evidence.
    To this end, we encouraged them to write queries which also matched a
    large number of items on the entire set of tables, and that did not
excessively overlap. This resulted in 39 queries for labeling 98,570 tables with
the corresponding type and 55 queries for labeling 165,302 columns.}

\leanparagraph{Gold standards} \conditional{To test the performance, the same
human annotators as before manually annotated 400 random tables.  The tables in
this sample have, on average,  9.92 rows (SD 7.28) and 5.07 columns (SD 3.20).}
These tables were annotated with the number of header rows, and table and column
types. This process resulted in 321 table type and 873 column type annotations
(excluding $\mathbf{Other}$). \conditional{ Most tables were annotated with the
    $\mathbf{Observation}$ class (258), followed by $\mathbf{Input}$ (50); the
    smallest class was $\mathbf{Example}$ (13).  The human annotators have
    annotated the table and column types looking at the images of the tables,
    the table captions, and possibly the full paper in case it was still not
    clear. The annotators have annotated the tables independently and resolved
    the conflicts together whenever they  disagreed. After the first round of
    annotation using the first version of the ontology (44 classes), we
    marked as infrequent all classes with fewer than 10 annotations. These
classes were removed from the ontology and the annotations were redirected to
$\mathbf{Other}$.} For the Tablepedia dataset, we used the annotations provided
by the original authors.

\conditional{We highlight two aspects of our gold standard that have a direct impact on the
    evaluation. First, in contrast to~\cite{tables_www2020}, we decided not to
    filter out tables that were incorrectly extracted by Tabula. This makes our corpus more challenging because it
    might contain errors due to incorrect parsing. Second, our choice of
    merging infrequent column types into the type $\mathbf{Other}$
    ensures that for each type
    there is always some evidence, but it has the downside that some classes in
    the long tail are ignored. Interpreting such types is an additional
challenge that deserves a thorough study in future work.}




\begin{figure}[t]
    \scriptsize
    \centering
    \subfloat[Header detection]{\raisebox{1.5em}{\label{tab:header-detect}\begin{tabular}{lc}
    Method &  Acc. \\
    \hline
    $1^{st}$ Row               &  0.71 \\
    Ours &  \bf{0.76} \\
\end{tabular}

}}
    \subfloat[Table type prediction on our corpus]{\raisebox{1em}{\label{tab:table-types}\begin{tabular}{lcccc}
Model &  Prec. &  Recall &   F1 &  AUC \\
\hline
SVM                 &            0.71 &         0.79 & 0.74 & 0.86 \\
LR &            0.72 &         0.79 & 0.74 & 0.84 \\
NB         &            \bf{0.80} &         \bf{0.82} & \bf{0.79} & \bf{0.91} \\

\end{tabular}
}}
    \subfloat[MV vs. Snorkel]{\label{tab:labelmodel}\begin{tabular}{lcc}

    Task &  MV & Snorkel \\
    \hline
    Table Types & 0.50 & \bf 0.71 \\
    Column Types & \multirow{2}{*}{\bf 0.56} & \multirow{2}{*}{0.49} \\
    (Our corpus) & & \\
    Column Types & \multirow{2}{*}{0.39} & \multirow{2}{*}{\bf 0.65} \\
    (Tablepedia) & & \\

\end{tabular}
}\\
 %
    \subfloat[Column type prediction on our corpus]{\raisebox{0.5em}{\label{tab:column-props}\begin{tabular}{lcccc}

Model &  Prec. &  Recall &   F1 &  AUC \\
\hline
NB         &  0.52 &   0.48 & 0.47 & \textbf{0.87} \\
SVM                 &  \bf{0.58} &   \bf{0.56} & \bf{0.53} & 0.83 \\
LR &  \bf{0.58} &   \bf{0.56} & \bf{0.53} & \bf{0.85} \\

\end{tabular}

}}
    \subfloat[Column type prediction on Tablepedia]{\label{tab:tablepedia}\begin{tabular}{lcccc}

Model &  Prec. &  Recall &   F1 &  AUC \\
\hline
Yu et al. \cite{tables_www2020} &0.82 &         0.81 & 0.81 & 0.90 \\
NB         &            0.84 &         0.82 & 0.81 & 0.96 \\
SVM                 &            0.90 &         0.89 & 0.89 & 0.97 \\
LR &            \bf{0.92} &         \bf{0.91} & \bf{0.91} & \bf{0.98} \\

\end{tabular}
}\\

    \caption{Table interpretation with Na\"ive Bayes (NB),
    Support Vector Machine (SVM), Logistic Regression (LR). MV is Majority
Voting, AUC is area under the curve}
    \label{fig:inter}
\end{figure}


\subsection{Table Interpretation}


Figure~\ref{tab:header-detect} reports the accuracy of our header detection
heuristic compared to the baseline that consists of always selecting the $1^{st}$ row. We
observe that our technique has superior performance, although it still makes
some mistakes.

In Figure~\ref{tab:table-types}, we report the performance of our table type
detection models on our gold standard. In general, we observe that all three
models return reasonably high performance. Na\"ive Bayes (NB) outperformed the others, especially in
terms of $F_1$ and AUC. Thus, we decided to select this as the default one for
this task.




In Figure~\ref{tab:column-props}, we report the classifiers' performance for the
column types on our gold standard, while Figure~\ref{tab:tablepedia} reports the
same for Tablepedia. In both cases, we see that LR performs best, likely due to
the combined importance of textual and numeric features for this task.
Additionally, we observe that our model significantly outperforms the model
of~\cite{tables_www2020} on their dataset. If we compare the scores between the
two datasets, then we see that they are significantly lower with our
dataset. The reason is two-fold: First, the authors of Tablepedia have manually
removed much noise from the extracted tables while no pre-processing took place on
our dataset. Second, our dataset contains many more classes than Tablepedia,
which makes it more challenging to predict.


Finally, we studied the added value of using Snorkel and compared it with a
simpler majority voting (MV), i.e., labeling a data point using the
most frequently predicted class. In Figure~\ref{tab:labelmodel}, we report both
the accuracy obtained with majority voting and with Snorkel with various types
of predictions. While Snorkel outperforms MV for the table type detection and
column type detection in Tablepedia, MV is better when detecting the column
types of our corpus. This was expected because, in this last case, our
labeling functions (i.e., SPARQL queries) have frequently abstained.
Consequently, $M$ has a low label density, and whenever this occurs,
Snorkel is unable to compute optimal weights that diverge from MV~\cite{snorkel}.

\subsection{Entity Linking}

\begin{figure}[t]
    \centering
    \scriptsize
    \subfloat[Ablation study. The bar marked with $r_i$ reports the number of
    entities when only EGD $r_i$
    is included in the rule set]{\label{fig:ablation}\includegraphics[width=0.7\textwidth]{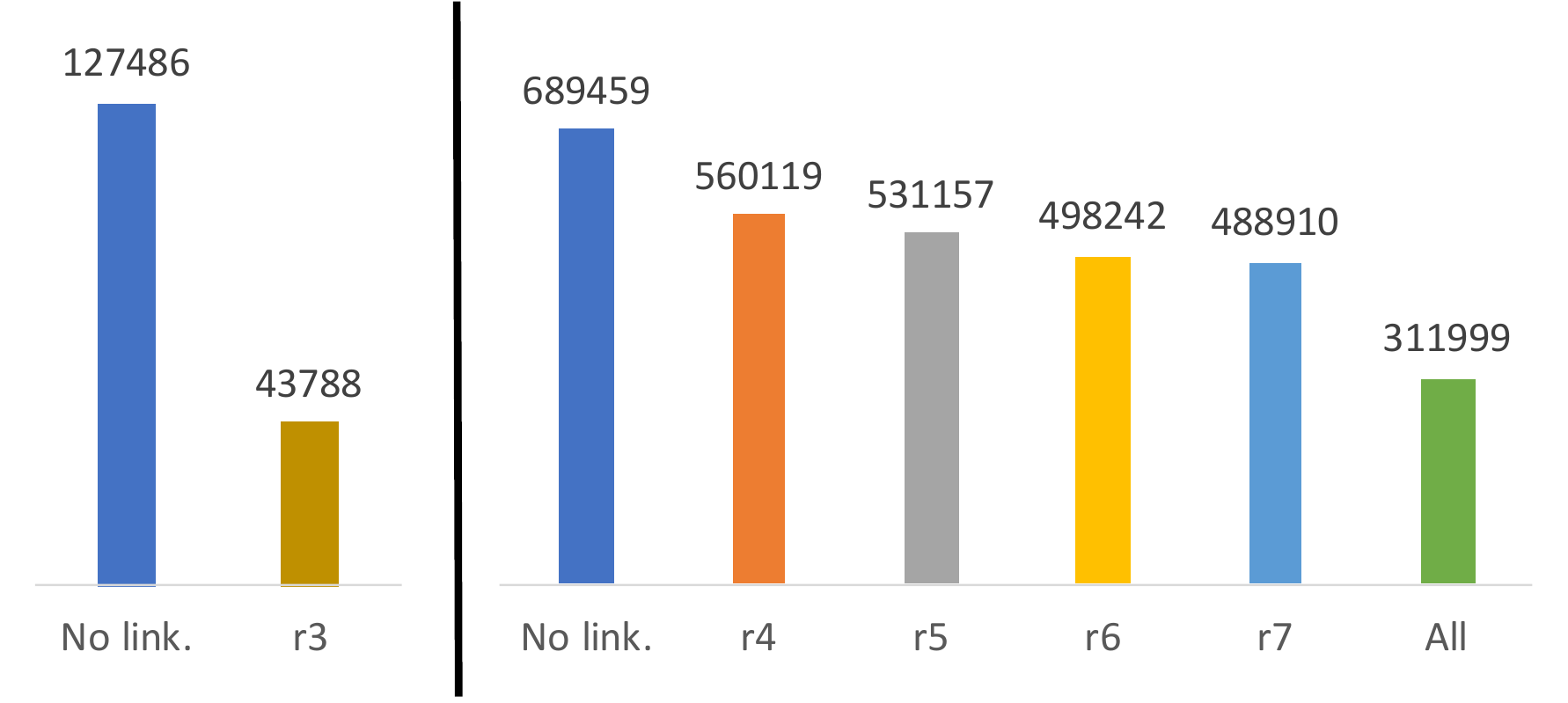}}
    \subfloat[Examples]{\raisebox{6em}{\label{fig:example}\definecolor{darkgreen}{rgb}{0.0, 0.5, 0.0}

\begin{tabular}{llcll}
\multirow{7}{*}{\rotatebox{90}{\textbf{Good} \textcolor{darkgreen}{\cmark}  }}
 & \textbf{Label} & \textbf{\# Links} \\
\hline
 & mnist  & 288 \\
 & knn   & 211 \\
 & wiki  & 146 \\
 & cifar-10 & 108 \\
 & en-es & 65 \\
\hline
\multirow{5}{*}{\rotatebox{90}{\textbf{Bad} \textcolor{red}{\xmark}}}
 & after & 183\\
 & analysis & 66 \\
 & subset & 49 \\
 & 0/0/0 & 9\\
 & f4(x) & 6 \\
\end{tabular}

}}
    \caption{Analysis of the performance of entity linking}
    \label{fig:entitylink}
\end{figure}

Figure~\ref{fig:ablation} reports the number of entities before and after the
execution of the EGD rules. The left side compares the number of entities that
refer to columns before and after $r_3$ was executed. As we can see, $r_3$
merged many entities, and this reduced the number of distinct entities of 65\%.
The right side shows the decrease of entities that refer to cells after the
execution of rules $r_4,\ldots,r_7$. Here, the bar titled $r_i$ reports the
number of entities if only $r_i$ is executed while the right-most column
indicates the number of entities when all rules are included. We observe that every
EGD contributes to merge some entities, but the best results are obtained when
all EGDs are activated: here, the EGDs merged about 55\% of the entities.

To evaluate the quality of entity links, we manually evaluated a sample of 100
merged entities. For each sampled entity, we first determined whether the entity
was a meaningful one. From this analysis, we discovered that 65\% of the
entities are correct while the remaining have either nonsensical labels or some
text resulted from errors of Tabula. In Figure~\ref{fig:example}, we report
examples of good and bad entities with their number of links.

Then, we looked at the cells which referred to the entity, which were 541 in
total. Since the rules could make a mistake and link two cells to the same
entity although they meant different ones, we evaluated, for each entity, the
precision of its links. Given the set of $n$ cells that link to the same entity,
the precision is computed by taking the cardinality of the largest subset of
cells that refer to the same concept and divide it by $n$. For instance,
consider an entity $X$ with label $Y$ which is linked to $n=4$ cells. Three of
these cells contain the text $Y$ but refer to a dataset while one cell contains
$Y$ but refers to something else. In this case, the precision for $X$ is
$\frac{3}{4}$. In our sample, the average precision over the meaningful entities
was about 97\%, which is a relatively high value. This indicates that reasoning
produced an accurate entity linking.

\section{Conclusion}
\label{sec:conclusion}

\leanparagraph{Summary} We presented \sys{}, an end-to-end system for building a
KB from the knowledge in scientific tables. One distinctive feature of \sys{} is
the usage of SPARQL queries for weak supervision to counter the lack of training
data. Another distinctive feature is the usage of existentially quantified rules
to link the entities without the help of a pre-existing KB.

Our pipeline effectively combines statistical-based classification and logical
reasoning, exploiting SPARQL and remote KBs like Semantic Scholar.  Therefore,
we believe that ours is an excellent example of how semantic web technologies,
statistical- and logic-based AI can be used side-by-side.

\leanparagraph{Future work} Although our results are encouraging, and the
current KB is already able to answer some non-trivial queries, future work is
required to improve the performance.  First, a more accurate table extraction
procedure is needed to improve the accuracy of table interpretation and entity
linking.  Moreover, our current ontology links classes only via $\sqsubseteq$. It
is interesting to study whether new relations can lead to better extractions.
For instance, specifying the range of some classes could be used to exclude
mappings to columns with incompatible values.  Finally, a natural continuation
of our work is to further research whether additional rules can return a better
entity linking. In particular, we believe that rules that take into account the
context of the table or co-authorship networks will be particularly useful.

To conclude, we believe that \sys{} represents one more step that brings us
closer to solve the problem of constructing an extensive and accurate KB of
scientific knowledge. Such a KB is a useful asset for assisting the researchers,
and it can play a crucial role in turning the vision of open science into a
reality.

\bibliography{references}
\bibliographystyle{splncs04}

\end{document}